\documentclass[lettersize,journal]{IEEEtran}
\usepackage{amsmath,amsfonts}
\usepackage{algorithmic}
\usepackage{algorithm}
\usepackage{array}
\usepackage[caption=false,font=normalsize,labelfont=sf,textfont=sf]{subfig}
\usepackage{textcomp}
\usepackage{stfloats}
\usepackage{url}
\usepackage{verbatim}
\usepackage{graphicx}
\usepackage{cite}
\usepackage{booktabs} %
\usepackage{multirow} 
\newcolumntype{L}[1]{>{\raggedright\let\newline\\\arraybackslash\hspace{0pt}}m{#1}}
\newcolumntype{C}[1]{>{\centering\let\newline\\\arraybackslash\hspace{0pt}}m{#1}}
\newcolumntype{R}[1]{>{\raggedleft\let\newline\\\arraybackslash\hspace{0pt}}m{#1}}

\begin{document}

\title{Enhancing Large Language Models with Reward-guided Tree Search for Knowledge Graph Question and Answering}

\author{Xiao Long, Liansheng Zhuang,~\IEEEmembership{Member, ~IEEE}, Chen Shen, Shaotian Yan, Yifei Li, Shafei Wang 
\thanks{Xiao Long, Liansheng Zhuang, and Yifei Li are with the
School of Cyberspace Science and Technology, University of Science
and Technology of China (USTC), Hefei, Anhui 230026, China (e-mail:
longxiao1997@mail.ustc.edu.cn; lszhuang@ustc.edu.cn).

Chen Shen and Shaotian Yan are independent researchers.

Shafei Wang is with the Peng Cheng Laboratory, Shenzhen 518066, China.}
}

\markboth{Journal of \LaTeX\ Class Files,~Vol.~14, No.~8, August~2021}%
{Shell \MakeLowercase{\textit{et al.}}: A Sample Article Using IEEEtran.cls for IEEE Journals}


\maketitle

\begin{abstract}
Recently, large language models (LLMs) have demonstrated impressive performance in Knowledge Graph Question Answering (KGQA) tasks, which aim to find answers based on knowledge graphs (KGs) for natural language questions. 
Existing LLMs-based KGQA methods typically follow the Graph Retrieval-Augmented Generation (GraphRAG) paradigm, which first retrieves reasoning paths from the large KGs, and then generates the answers based on them. 
However, these methods emphasize the exploration of new optimal reasoning paths in KGs while ignoring the exploitation of historical reasoning paths, which may lead to sub-optimal reasoning paths. Additionally, the complex semantics contained in questions may lead to the retrieval of inaccurate reasoning paths.
To address these issues, this paper proposes a novel and training-free framework for KGQA tasks called Reward-guided Tree Search on Graph (RTSoG). RTSoG decomposes an original question into a series of simpler and well-defined sub-questions to handle the complex semantics. 
Then, a Self-Critic Monte Carlo Tree Search (SC-MCTS) guided by a reward model is introduced to iteratively retrieve weighted reasoning paths as contextual knowledge. 
Finally, it stacks the weighted reasoning paths according to their weights to generate the final answers. Extensive experiments on four datasets demonstrate the effectiveness of RTSoG. Notably, it achieves 8.7\% and 7.0\% performance improvement over the state-of-the-art method on the GrailQA and the WebQSP respectively.
\end{abstract}


\section{Introduction}
Knowledge Graph Question Answering (KGQA) has drawn much attention in recent years. It aims to find answers for natural language questions based on the knowledge graphs (KGs), such as Freebase~\cite{bollacker2008freebase}, Wikidata~\cite{vrandevcic2014wikidata}, and DBpedia~\cite{auer2007dbpedia}, which are built from numerous triplets consisting of (head entity, relation, tail entity). Since natural language questions often contain compositional semantics~\cite{lan2022complex}, accurately understanding the semantic information in the questions and identifying the structured knowledge in KGs is very important for KGQA tasks. 

Recently, as large language models (LLMs)~\cite{4o-mini, chatgpt, yang2024qwen2} have demonstrated significant progress in solving various complex NLP tasks~\cite{wei2022chain, yao2024tree}, some works begin to exploit the potential of LLMs in solving the task of knowledge graph question answering~\cite{markowitz2024tree, ma2024think}. Typically, existing LLM-based methods~\cite{chen2024plan, sun2023think} follow the paradigm of Graph Retrieval-Augmented Generation (GraphRAG)~\cite{peng2024graph}, which consists of the Graph-guided retrieval stage and the Graph-enhanced generation stage. In the first stage, they retrieve reasoning paths in KGs by different search strategies. In the second stage, they utilize the retrieved reasoning paths as contextual knowledge for LLMs to generate the answers to the questions. Under this paradigm, LLMs are often training-free and the quality of retrieval reasoning paths plays a key role in the final performance. Various methods are different in the search strategies as shown in Table~\ref{t1}. Benefiting from the great reasoning abilities of LLMs, GraphRAG-like methods have shown impressive performance in the task of knowledge graph question answering. 
\begin{table}[h]
\centering
\caption{Comparison of different methods in Graph-guided retrieval methods.}
\label{t1}
\begin{tabular}{cc}
\toprule
\textbf{Algorithms} & \textbf{Graph-guided retrieval methods} \\ \midrule
ToG~\cite{sun2023think}             & Beam search                                \\
ToT~\cite{markowitz2024tree}             & Greedy search                 \\
PoG~\cite{chen2024plan}             & Best-of-N                      \\
ToG2.0~\cite{ma2024think}          & Beam search                   \\ \midrule
RTSoG         & SC-MCTS                        \\ \bottomrule
\end{tabular}
\end{table}

However, existing LLM-based KGQA methods still suffer from the following issues.
Firstly, in the graph-guided retrieval stage, most existing methods emphasize the exploration of new optimal reasoning paths in KGs while ignoring the exploitation of existing reasoning paths. This may lead to finding sub-optimal reasoning paths. 
Secondly, existing methods often use the original question to guide the retrieval. Since the questions often contain compositional semantics~\cite{lan2022complex}, this may lead to retrieving inaccurate reasoning paths from KGs. 
Finally, in the graph-enhanced generation stage, existing methods usually treat the different retrieval reasoning paths equally to reason the answer, ignoring the differences in retrieved reasoning paths. These issues significantly limit the performance of Large Language Models under the existing GraphRAG paradigm.

Inspired by the above insights, this paper proposes a novel framework named Reward-guided Tree Search on Graph (RTSoG) for KGQA tasks. Its core idea is to use the Monte Carlo Tree Search (MCTS) to better balance the exploration and exploitation of reasoning paths in KGs. Specifically, the proposed RTSoG consists of three stages: Question Decomposition, Weighted Reasoning Paths Retrieval and Answer Generation. In the question decomposition stage, it leverages LLMs to decompose a complex question into a series of simpler and more clearly defined sub-questions, which along with the original question, serve as guidance for the subsequent stages. Next, it iteratively performs the Self-Critic Monte Carlo Tree Search (SC-MCTS) on the KGs to retrieve weighted reasoning paths that support the inference of the question. The self-critic mechanism provides timely termination signals during the iteration, which can avoid unnecessary exploration. Finally, to consider the differences brought by reasoning paths with varying weights, we propose a reasoning path stack to assist in generating the final answer. To summarize, our contributions are as follows:
\begin{itemize}
    \item A novel Reward-guided Tree Search on Graph (RTSoG) is proposed for KGQA tasks. It extracts more accurate contextual knowledge from KGs via a variant of MCTS to enhance the reasoning capabilities of LLMs, and thus achieves SOTA performance against existing KGQA methods. 
    \item A Self-Critic Monte Carlo Tree Search (MCTS) guided by a reward model is introduced to retrieve the weighted reasoning paths in KGs as exact contextual knowledge for LLMs to answer the questions. The Self-Critic mechanism can significantly enhance both efficiency and performance in identifying optimal reasoning paths.
    \item Extensive experiments on four benchmark datasets demonstrate that RTSoG achieves superior performances in KGQA tasks. Notably, RTSoG achieves 8.7\% and 7.0\% improvement over all types of state-of-the-art methods on the GrailQA and the WebQSP respectively.
\end{itemize}

\section{Related Wrok}
In this section, we introduce existing knowledge graph question and answering methods and large language model reasoning methods, then we explain their relation to our work.

\textbf{Knowledge Graph Question and Answering.} Knowledge graph question answering aims to find answer entities that are multiple hops away from the topic entities in a large-scale KG. There are two mainstream approaches to solve the KGQA task: Semantic parsing-based methods and Information retrieval-based methods.
Semantic parsing-based methods focus on translating questions into logical forms executable against KGs, such as SPARQL, query graph, and S-expression. Some early methods rely on pre-defined question templates to synthesize queries~\cite{reddy2014large}, which requires strong domain knowledge. Recent methods like StructGPT~\cite{jiang2023structgpt} utilize strategies of step-wise query graph generation and search for semantic parsing.
Alternatively, other SP-based methods, like RnG-KBQA~\cite{ye2021rng} and ArcaneQA~\cite{gu2022arcaneqa} employ sequence-to-sequence models to generate S-expressions completely and offer various enhancements to the semantic parsing process. Some other works~\cite{ren2020query2box, long2022neural} use the FOL queries to help the parsing process. However, this type of method needs ground-truth executable queries as supervision (which are costly to obtain) and their performance is limited when the KG has missing links (non-executable queries).

Information retrieval-based methods for KGQA retrieve reasoning paths~\cite{zhang2022subgraph}, which are used as input during KGQA reasoning. The earlier IR methods~\cite{he2021improving} use neural networks to directly score candidate answers and determine an answer set based on a score threshold. Subsequently, some works utilize Graph Neural Networks (GNNs) for information propagation and score the results accordingly~\cite{sun2019pullnet}. They also called Graph-based retrieval methods~\cite{mavromatis2024gnn}. Recently, to integrate Large Language Models(LLMs) for KGQA, retrieval-augmented methods leverage LLMs to retrieve the relative facts from the KGs to improve the reasoning performance~\cite{chen2024plan, sun2023think, luo2023reasoning, long2025eperm}. UniKGQA~\cite{jiang2022unikgqa} unifies the graph retrieval and reasoning process into a single model with LLMs. ToG~\cite{sun2023think} uses LLMs to iteratively execute beam search on the KGs, discovers the most promising reasoning paths, and returns the most likely reasoning results. PoG~\cite{chen2024plan} utilizes LLMs to perform adaptive planning on Knowledge Graphs, which can be regarded as a Best-of-N (BoN) search on the KGs. ToT~\cite{markowitz2024tree} leverages LLMs to iteratively execute a greedy search on the KGs and then generate the answers. However, these methods emphasize the exploration of new optimal reasoning paths in KGs while ignoring the exploitation of existing reasoning paths. This may lead to finding sub-optimal reasoning paths.

\textbf{Large Language Model Reasoning.} LLMs have shown significant advantages in semantic understanding, generation, and reasoning. Previous studies, such as Chain-of-thoughts(CoT)~\cite{wei2022chain} imitates the thought process of humans to provide step-by-step solutions given a question. Self-Consistency CoT~\cite{wang2022self} improves the reliability and Self-Consistency of answers by sampling multiple interpretations from LM and then selecting the final answer that appears most frequently. Tree-of-Thoughts (ToT)~\cite{yao2023tree} further generalizes the CoT methodology by considering multiple different reasoning paths in the tree and exploring coherent units of thought to execute thoughtful decision-making. And ReAct~\cite{yao2023react} introduces a prompt-based agent that uses the large language model to interact with the environment.
Alternatively, other studies which on top of LLMs has recently attracted significant attention. They have explored search algorithms to improve the performance of LLMs during the inference stage~\cite{wang2024q, qi2024mutual, zhang2024llama}. Many studies have proven that scaling the inference-time computation can lead to substantial improvements in the performance of LLMs without training~\cite{brown2024large, snell2024scaling, zhang2024rest}, which sampling diverse reasoning paths can significantly enhance the probability of finding the correct answers. 
Meanwhile, these search algorithms, where multiple branches of outcomes are explored during the search, have been widely applied in reinforcement learning algorithms~\cite{silver2017mastering} and many real-world applications, such as AlphaGo~\cite{silver2016mastering} and MuZero~\cite{schrittwieser2020mastering} for their good exploration-exploitation trade-off. However, current enhancing LLMs with search approaches mainly rely on the internal knowledge of LLMs to search potential solutions, which will encounter the issues of hallucination and out-of-date knowledge in large language models and lead to a decline in model performance. 
In this paper, we leverage the knowledge retrieved from the more reliable knowledge graphs to enhance the faithful reasoning ability of the large language model.

\section{Preliminary}

\textbf{Knowledge Graph (KG)} stores massive factual knowledge in the form of a set of triplets: $G=\{(e, r, e') \mid e, e' \in E, r \in R \}$, where $E$ and $R$ denote the set of entities and relations, respectively.

\textbf{Knowledge Graph Question Answering (KGQA)} is a typical reasoning task based on KGs. Given a natural language question \( q \) and the KG \( \mathcal{G} \), the task aims to design a function \( f \) to predict answer entities \( e_a \in \mathbb{A}_q \) based on knowledge from \( \mathcal{G} \), i.e., \( e_a = f(q, \mathcal{G}) \). Following previous works~\cite{jiang2022unikgqa, sun2019pullnet}, we assume the topic entities \( e_{t} \in \mathbb{T}_q \) mentioned in the question \( q \) and answer entities \( e_a \in \mathbb{A}_q \) are labeled and linked to the corresponding entities in knowledge graph \( \mathcal{G} \), i.e., \( \mathbb{T}_q, \mathbb{A}_q \subseteq \mathcal{E} \).

\textbf{Realation Paths} are a sequence of relations: $z = \{ r_1, r_2, \ldots, r_l \}$, where $r_i \in R$ denotes the $i$-th relation in the path and $l$ denotes the length of the path.

\textbf{Reasoning Paths} are the instances of a relation path $z$ from a topic entity to target entity in the KG, e.g., the reasoning path $P_{z_l}$ from $e_0 \in \mathbb{T}_q$ to $e_l$: $P_{z_l} = e_0 \stackrel{r_1}{\longrightarrow} e_1 \stackrel{r_2}{\longrightarrow} \ldots \stackrel{r_l}{\longrightarrow} e_l$ where $e_i \in E$ denotes the $i$-th entity and $r_i$ denotes the $i$-th relation in the relation paths $z_l$.

\begin{figure*}[ht]
\centering
{\includegraphics[height=0.53\textwidth]{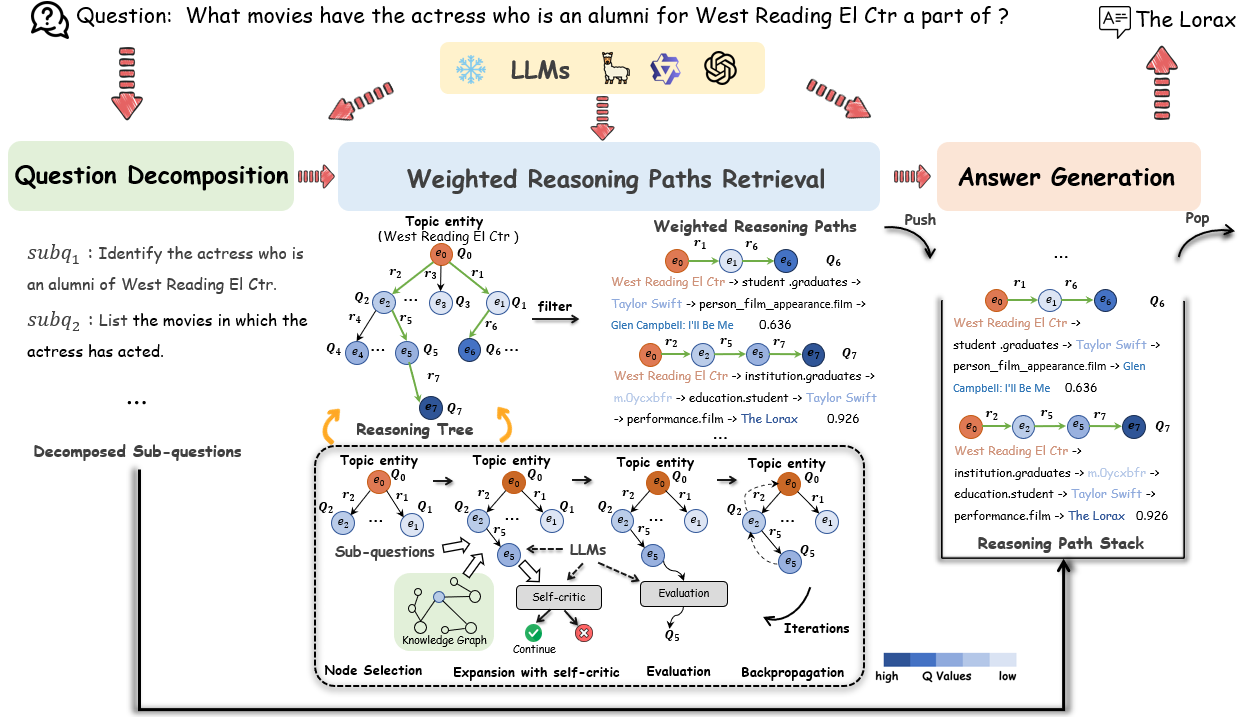}} 
\caption{Overview of the proposed training-free RTSoG framework, which contains the three stages: Question Decomposition, Weighted Reasoning Paths Retrieval and Answer Generation.}
\label{p1}
\end{figure*}

\section{Methodology}
In this section, we introduce the technical details of the novel RTSoG framework for KGQA tasks. As illustrated in Fig~\ref{p1}, RTSoG consists of three stages: Question Decomposition, Weighted Reasoning Paths Retrieval and Answer Generation. It should be noted that the based LLM used in the three stages is training-free and called the policy model $\pi_{\theta}$. Firstly, RTSoG utilizes the policy model $\pi_{\theta}$ to decompose the original question into a series of simpler sub-questions, which serve as guidance for the subsequent stages. In the second stage, RTSoG utilizes the policy model $\pi_{\theta}$ to iteratively perform SC-MCTS on the KG to explore reasoning paths guided by the reward model $V_{\theta}$, which is the same large language model with the policy model. Then, we can obtain a reasoning tree $\mathcal{T}$. In the next step, we filter the top $K$ paths with higher $Q$ values from the reasoning tree as weighted reasoning paths that support the inference of the question. Finally, we consider the differences brought by paths with different weights. Paths are pushed into the reasoning path stack $\mathcal{S}$ in descending order of their weights for final reasoning path determination. The higher-weight valid paths that enter the stack earlier serve as historical information to assist in judging subsequently pushed paths. Consequently, the remaining paths in the stack represent all correct historical paths, which the policy model $\pi_{\theta}$ ultimately uses as the basis for generating the final answer. Next, we provide a detailed description of the three stages in RTSoG.
\subsection{Question Decomposition}
To reduce the difficulty of understanding the semantics of the original question and enable more accurate retrieval of reasoning paths, RTSoG leverages the policy model $\pi_{\theta}$ to decompose the original question into a series of simpler and more precise sub-questions in the first stage. Specifically, we prompt the large language model to utilize both the question and its associated topic entities $e_0$ to break down the original question $q$ into a set of sub-questions. These sub-questions, denoted as $subq = \{ subq_1, subq_2, \cdots, subq_{n} \}$, are subsequently used for guiding knowledge exploration on the KG via SC-MCTS. It should be noted that the LLM performing question decomposition is also training-free.
\begin{equation}
   \{ subq_i \}_{i=1}^{n} = \pi_{\theta}(q, e_0).
\end{equation}
The decomposed sub-questions $\{ subq_i \}_{i=1}^{n}$ generally analyze the problem from different perspectives, which can better guide the subsequent stages: retrieval of weighted reasoning paths and the generation of answers.

\subsection{Weighted Reasoning Paths Retrieval}
In this stage, we first construct a reasoning tree $\mathcal{T}$ in the knowledge graph $\mathcal{G}$ and then filter the high-weighted reasoning paths related to the question $q$. To achieve this, we propose a Self-Critic Monte Carlo Tree Search (SC-MCTS). As illustrated in Fig~\ref{p1}, the SC-MCTS method combines the Monte Carlo Tree Search (MCTS) with the Self-Critic mechanism. It divides each iteration into four steps: node selection, expansion with self-critic, evaluation and backpropagation. Next, we will describe each step in detail and highlight the differences between SC-MCTS and traditional MCTS on the graph.

\textbf{Node Selection.} 
SC-MCTS builds a reasoning tree $\mathcal{T}$ based on the same training-free LLM: policy model $\pi_{\theta}$ and value model $V_\theta$. Each node $s_i=[e_i, P_{z_i} N(s_i), Q(s_i)]$ represents a state comprising the entity $e_i$ in KG, its historical reasoning path $P_{z_i}$ is from the topic entity $e_0 \in \mathbb{T}_q$ to current entity $e_i$ connected by the corresponding relations, the number of visits $N(s_i)$, and $Q(s_i)$, which denotes the reward value for inferring the answers to the question and its sub-questions based on the node's historical reasoning path $P_{z_i}$.
At each iteration, the algorithm first selects an appropriate node from the current tree for subsequent exploration and expansion. Specifically, starting from the root node $s_0$ (which always corresponds to the topic entity $e_0$), it traverses the tree by selecting a child node at each level until reaching a leaf node. A leaf node is defined when either of the following conditions is met: (1) reaching the maximum tree depth or (2) receiving an End of Search (EoS) signal. To balance the exploration and exploitation, we employ the well-known Upper Confidence Bounds applied to Trees (UCT)~\cite{kocsis2006bandit} for node selection:
\begin{equation}
    UCT(s_i) = \frac{Q(s_i)}{N(s_i)} + c \cdot \sqrt{\frac{\ln N(p)}{N(s_i)}},
    \label{eq2}
\end{equation}
Here, $Q(s_i)$ is the reward value for inferring the answers to the question and its sub-questions based on the node's historical reasoning path $P_{z_i}$ by value model. The $N(s_i)$ in equation~\ref{eq2} represents the number of visits to node $s_i$, $p$ is the parent node of $s_i$, and $c$ is the exploration weight. The node with the highest UCT value is selected for subsequent steps.

\textbf{Expansion with self-critic.}
After selecting the current node $s_i$, RTSoG utilizes the policy model $\pi_{\theta}$ to incorporate the corresponding reasoning path $P_{z_i}$ as historical reasoning information. It then samples $k$ most relevant entities for solving both the main question and its sub-questions from the knowledge graph $\mathcal{G}$ and subsequently expands the search tree by adding these entities as new nodes. The equation~\ref{eq3} describes this process, and it involves two phases: relation exploration and entity exploration. We provide detailed explanations in subsequent sections.
\begin{equation}
    s_j=Get\_child(subq, s_{i}, \mathcal{G}, \pi_{\theta}).
    \label{eq3}
\end{equation}

\textit{(1) Relation Exploration.} Relation exploration is a process to retrieve all adjacent relations $R^{e_i}_{adj}$ of the entity $e_i$ corresponding to the current node $s_i$, and filter out the relations most relevant to the question $q$ and sub-questions $subq$. As equation~\ref{eq4}, we first conduct the predefined relation search to obtain all relations in the known KG $\mathcal{G}$ linked to the selected entity $e_i$ as the candidate relation set $R^{e_i}_{adj} = \{ r^{e_i}_{adj, 1}, r^{e_i}_{adj, 1}, \cdots, r^{e_i}_{adj, N} \}$, where $N$ is the total number of adjacent relations. RTSoG then employs the value model $V_{\theta}$ to perform adjacent relation filtering: based on the semantic information of sub-questions, topic entity information, and current historical reasoning path, it dynamically selects a variable number $b$ of relevant adjacent relations $R^{e_i}_{filt}=\{r^{e_i}_{filt, 1}, \cdots, r^{e_i}_{filt, b}\}$ from $R^{e_i}_{adj}$, and assigns reward scores $S^{r}_{filt}=\{s^{r}_{filt, 1}, \cdots, s^{r}_{filt, b} \}$ that reflect their likelihood of solving the sub-questions. These filtered relations and their reward will serve as known information to assist in subsequent entity exploration and determine the entities to be expanded on the tree.
\begin{equation}
    R^{e_i}_{adj}=Get\_adj\_relations(s_{i}, \mathcal{G}),
    \label{eq4}
\end{equation}
\begin{equation}
    R^{e_i}_{filt}, S^{r}_{filt} = V_{\theta}(s_{i}, R^{e_i}_{adj}, \mathcal{G}).
    \label{eq5}
\end{equation}
\textit{(2) Entity Exploration.} Due to the multiple semantics of relations~\cite{lin2015learning, long2024kgdm, long2024fact} in the knowledge graph (KG), there are multiple tail entities given a known head entity and the filtered relations. Therefore, we should filter the most reasoning-related entities as the final expanded nodes for every $b$ relation. Given the entity $e_i$ corresponding to the current node $s_i$ and a filtered relation $r^{e_i}_{filt, l}$ in $R^{e_i}_{filt}$, the model performs a predefined query operation on the KG to obtain the set of all candidate tail entities $E^{e_i, r_{filt, l}}_{cand}=\{ e^{e_i, r_{filt, l}}_{cand, 1}, \cdots, e^{e_i, r_{filt, l}}_{cand, M} \}$ in equation~\ref{eq6}, where $M$ is the total number of adjacent entities. Then, we use the entity $e^{e_i, r_{filt, l}}_{cand, j}$ in $E^{e_i, r_{filt, l}}_{cand}$ as candidate entities and combine them with their preceding relations to extend the current reasoning path, forming the candidate reasoning paths $P_{cand} = \{P_{cand, j} \}^{M}_{j=1}$ where $P_{cand, j} = P_{e_i} \stackrel{r^{e_i}_{filt, l}}{\longrightarrow} e^{e_i, r_{filt, l}}_{cand, j}$. Next, in equation~\ref{eq7}, we prompt the value model $V_\theta$ to evaluate each candidate historical reasoning path to obtain the reward, determining the likelihood of successfully inferring the sub-questions through that reasoning path.
\begin{equation}
    E^{e_i, r_{filt, l}}_{cand} = Get\_tail\_entities(s_{i}, r^{e_i}_{filt, l}, \mathcal{G}),
    \label{eq6}
\end{equation}
\begin{equation}
    S^{P}_{cand} = V_{\theta}(subq, e_0, P_{cand}),
    \label{eq7}
\end{equation}
where $S^{P}_{cand}=\{ s^{P}_{cand, j} \}^{M}_{j=1}$. We select the index of highest reward candidate reasoning path $j_{\text{max}} = \arg\max_{j}(s^{P}_{cand, j})$ and extend its corresponding entity $e^{e_i, r_{filt, l}}_{cand, j_{max}}$ and their preceding relation $r^{e_i}_{filt, l}$ as new child node $e_{i+l}$ of the $e_i$ on the tree for the next iterative exploration. For every relation that passes the filtering process, the highest-reward entity is chosen as the final node for expansion.

Finally, unlike traditional MCTS, which proceeds to the rollout phase immediately after node expansion, SC-MCTS performs instant evaluation right after expanding child nodes. Although the value model can assess historical reasoning paths, it cannot generate a signal to terminate node expansion. Specifically, even after reaching a correct node, the algorithm may continue exploring further beneath it, leading to reduced search efficiency and the generation of incorrect reasoning paths. To address this issue, we propose a self-critic mechanism to provide timely termination signals, thereby avoiding unnecessary exploration and guiding deeper search.  In practice, after each node expansion, we prompt the policy model $\pi_{\theta}$ to generate an End-of-Search (EoS) signal based on whether the currently expanded historical reasoning path $S^{P}_{cand, j_{max}}$ can infer the answer to the question in equation~\ref{eq8}. If an EoS signal is received, the node is set as a leaf node and will not be expanded in subsequent iterations. 
\begin{equation}
    EoS(s_j) = \pi_{\theta}(subq, s_j).
    \label{eq8}
\end{equation}

\textbf{Evaluation.} Traditional MCTS methods require performing lots of rollouts from the current node until the task ends to evaluate the expanded nodes. While in our work, we use the LLM as the value model $V_\theta$~\cite{lightman2023let} to evaluate the current node. It assigns an estimated reward to the expanded node $e_{i+l}$, which effectively quantifies the effectiveness of the policy model $\pi_{\theta}$ in successfully answering the input question based on the historical reasoning path of the current node. The evaluation involves two parts: local expansion relation assessment $S^{r}_{filt}$ and global historical reasoning path assessment $S^{P}_{cand}$, as mentioned in the relation exploration and entity exploration above. The total reward value of $e_{i+l}$ is a weighted combination of global and local reward in equation~\ref{eq9}, where $\alpha$ controls the relative influence of each part.
\begin{equation}
    Q_{s_{i+l}} = \alpha s^{r}_{filt, l} + (1-\alpha) s^{P}_{cand, l}.
    \label{eq9}
\end{equation}

\textbf{Backpropagation.} After obtaining the final reward for the newly expanded node, we conduct value backpropagation starting from the new node $s_{i+l}$. The value of every parent node of $s_{i+l}$ is updated using a weighted average method. For every node $C$ on the trace from root to $s_{i+l}$, we update its $N_{C}$ and $Q_C$ as follows:
\begin{equation}
    N(C) \leftarrow N(C) + 1,
    \label{eq10}
\end{equation}
and 
\begin{equation}
    Q(C) \leftarrow \frac{\sum_{i} n_{C_i} \cdot v_{C_i}}{\sum_{i} n_{C_i}},
    \label{eq11}
\end{equation}
where $C_i$($i=1, 2, \cdots, k$) are the children of $C$. This actually updates the value of $C$ according to its children's value expectation.

We iteratively repeat the above four steps to explore new entities and their relations in KGs as new nodes in the reasoning tree, completing $H$ rounds of iterations to obtain the final reasoning tree $\mathcal{T}$, where $H$ is a hyper-parameter, and we will analyze it in experiments. Finally, we select the Top-$K$ nodes $\{e_{rng, i}\}^{K}_{i=1}$ with higher reward value $Q$ from the reasoning tree and use their corresponding historical reasoning paths $\{P_{rng, i}\}^{K}_{i=1}$ as weighted reasoning paths that support to generate the final answer in the final stage.
\begin{equation}
    \{P_{rng, i}\}^{K}_{i=1} = Get\_weighted\_reasoning\_paths(\mathcal{T}_{q})
    \label{eq12}
\end{equation}
\begin{equation}
    e_a = \pi_{\theta}(\mathcal{S}, q, subq)
    \label{eq13}
\end{equation}

\subsection{Answer Generation}
In the final stage, we utilize the filtered $K$ weighted reasoning paths $\{P_{rng, i}\}^{K}_{i=1}$ as the final information for generating the answer to the question. It is important to note that, unlike existing methods that directly generate answers based on retrieved relevant information, we consider the varying impacts of different weighted paths during reasoning in this stage. Specifically, we sort the $K$ weighted reasoning paths in descending order of their weights and introduce the reasoning path stack $\mathcal{S}$. Each time, the reasoning path with the highest weight is pushed into the stack, and the policy model $\pi_{\theta}$ is prompted to determine whether this path can derive the answer to the original question or its sub-questions. If the output is affirmative, the path is stored in the stack as known reasoning information to assist in the judgment of subsequent weighted reasoning paths. Finally, as equation~\ref{eq13}, RTSoG uses the policy model $\pi_{\theta}$ to generate the final answer $e_a$ based on the path remaining in the stack. The pseudocode of the RTSoG is shown in Algorithm~\ref{alg1}.

\begin{algorithm}
\caption{The proposed reward-guided tree search algorithm RTSoG.}
\label{alg1}
\begin{algorithmic}[1]
\STATE \textbf{Input:} KG $\mathcal{G}$, question $q$, topic entity $e_0$, iterations $H$, width of tree $b$, the number of subquestions $n$, the number of paths in $\mathcal{S}$, policy model \(\pi_{\theta}\), value model \( V_{\theta} \).
\STATE $ subq =  \{ subq_i \}_{i=1}^{n} \leftarrow \pi_{\theta}(q, e_0).$ \ 
\STATE \( \mathcal{T}_q \leftarrow \) Initialize\_tree(\( s_0 \))
\FOR{$h$ in range(\( H \))}
    \STATE \( s \leftarrow \) root(\( \mathcal{T}_q \))
    \STATE \rule{1cm}{0.4pt}  \textbf{\textit{Node Selection}}   \rule{1cm}{0.4pt}
    \WHILE{$s$ is not leaf node}
        \STATE \( s \leftarrow \) $argmax_{s^{'} \in children(s)} (\frac{Q(s^{'})}{N(s^{'})} + c \cdot \sqrt{\frac{\ln N(s)}{N(s^{'})}})$,
    \ENDWHILE    
    \STATE \rule{1cm}{0.4pt} \textbf{\textit{Expansion with Self-Critic}} \rule{1cm}{0.4pt}
        \FOR{$j$ in range(\( b \))}
            \STATE \( s_j \leftarrow \) $Get\_child$(\( subq, s, \mathcal{G}, \pi_{\theta} \)) \ //  \textit{Node expansion} 
            \STATE  \( EoS(s_j) \leftarrow \) $\pi_{\theta}$(\( subq, s_j \)) \ //  \textit{Self Critic}
            \STATE \rule{1cm}{0.4pt} \textbf{\textit{Evaluation}} \rule{1cm}{0.4pt}
            \STATE \( Q(s_j) \leftarrow V_{\theta}(subq, s_j, \mathcal{G}) \)
        \ENDFOR

    \STATE \rule{1cm}{0.4pt} \textbf{\textit{Value Backpropagation}} \rule{1cm}{0.4pt}
    \STATE Back\_propagate(\( s \)) \ // \textit{Update value of all nodes}
\ENDFOR  \ \ // \textit{Reasoning tree construction}
\STATE \( \{P_{rng, i}\}^{K}_{i=1} = \) $Get\_weighted\_reasoning\_paths$(\( \mathcal{T}_{q} \)) 
 \ 
\STATE \( \{P_{rng, m}\}^{K}_{m=1} = \) Sorted(\( \{P_{rng, i}\}^{K}_{i=1}\))
\STATE Reasoning\_path\_stack \( \mathcal{S} \leftarrow [ \ ] \)
\FOR{$m$ in range(\( K \))}
    \IF{\( \pi_{\theta}(\mathcal{S}, q, subq, P_{rng, m}) \)} 
        \STATE \( \mathcal{S} \leftarrow P_{rng, m} \)
    \ENDIF       
\ENDFOR
\STATE \( e_a = \pi_{\theta}(\mathcal{S}, q, subq) \) \ // \textit{Answer generation}
\STATE Return \( e_a \) 
\STATE \textbf{Output:} \( e_a \)
\end{algorithmic}
\end{algorithm}

\section{Experiments}
In this section, we comprehensively evaluate RTSoG by answering the following evaluation questions (EQs):
\begin{itemize}
\item \textbf{EQ1 (Main results).} Does RTSoG outperform other KBQA methods?
\item \textbf{EQ2 (Sensitivity analysis).} How sensitive is the model accuracy of RTSoG to the hyperparameters?
\item \textbf{EQ3 (Ablation study).} How does each of the proposed strategies in RTSoG contributes to the model accuracy?
\item \textbf{EQ4 (Efficiency analysis).} How does the RTSoG compare in efficiency to existing methods when inferring questions?
\item \textbf{EQ5 (Case study).} How does the RTSoG can find the reasoning paths to get the final answer?
\end{itemize}

\subsection{Experimental Setups}
\noindent \textbf{Datasets.} To demonstrate the effectiveness of RTSoG on knowledge graphs question answering, we adopt three representative multi-hop KGQA datasets: ComplexWebQuestions~\cite{talmor2018web}, WebQSP~\cite{yih2016value}, GrailQA~\cite{gu2021beyond} and WebQuestions~\cite{berant2013semantic}. The statistics of datasets are shown in Table~\ref{tab_a1}. WebQuestions is an open-domain QA dataset based on Freebase. WebQSP contains questions from WebQuestions that are answerable by Freebase. It tests I.I.D. generalization on questions. ComplexWebQuestions (CWQ) extends WebQSP and encompasses four types of complex questions: conjunction, composition, comparative, and superlative. GrailQA is a diverse KGQA dataset built on Freebase, and is designed to test three levels of generalization of models: I.I.D., compositional, and zero-shot. All four datasets rely on the external knowledge graph from Freebase~\cite{bollacker2008freebase}. For the large dataset GrailQA and WebQuestions we utilize the same testing samples as those in previous works~\cite{sun2023think, chen2024plan} to improve computational efficiency.

\begin{table}[h]
\centering
\caption{The statistics of the datasets used in this paper. * denotes we randomly selected 1,000 samples from the GrailQA and 1500 samples from WebQuestions test set as previous works.}
\begin{tabular}{lccc}
\toprule
\multicolumn{1}{c}{\textbf{Dataset}}     & \textbf{Answer Format} & \textbf{Train}              & \textbf{Test}              \\ \midrule
\multicolumn{1}{l|}{ComplexWebQuestions} & Entity                 & \multicolumn{1}{c|}{27,734} & \multicolumn{1}{c}{3,531}                \\
\multicolumn{1}{l|}{WebQSP}              & Entity/Number          & \multicolumn{1}{c|}{3,098}  & \multicolumn{1}{c}{1,639}        \\
\multicolumn{1}{l|}{GrailQA$^{*}$}             & Entity/Number          & \multicolumn{1}{c|}{44,337} & \multicolumn{1}{c}{1,000}                 \\
\multicolumn{1}{l|}{WebQuestions$^{*}$}        & Entity/Number          & \multicolumn{1}{c|}{3,778}  & \multicolumn{1}{c}{1,500}        \\ \bottomrule
\end{tabular}
\label{tab_a1}
\end{table}

\noindent \textbf{Evaluation Metrics.}Following prior research~\cite{li2023chain, baek2023knowledge}, we use exact match accuracy (EM) as the evaluation metric for all datasets.

\noindent \textbf{Baselines.} Due to variations in the performance of the method across different datasets, we select prior state-of-the-art (SOTA) approaches as baselines for each dataset. They can be categorized into three groups: 

(1) LLM-only methods, including Qwen2.5-14B~\cite{yang2024qwen2}, a powerful, open-source, large-scale language model (LLM) that has been pre-trained on a diverse range of tasks. Llama-3.1-8B~\cite{llama3.1}, is the updated version of Llama-2 with more powerful reasoning capabilities. ChatGPT~\cite{chatgpt}, is a powerful closed-source LLM that could follow instructions to conduct complex tasks. GPT4~\cite{gpt4}, is the new flagship model of OpenAI that could reason across different modalities and tasks. And GPT-4o-mini~\cite{4o-mini} is the mini-version of GPT-4o that could reason across different modalities and tasks. For LLM-Only Methods, we use chain-of-thought reasoning method~\cite{wei2022chain} that prompts LLMs to generate a chain of reasoning steps and the final answers. 

(2) Graph-based retrieval methods, including GraftNet~\cite{sun2018open}, a method retrieves relevant subgraphs from KGs with entity linking. PullNet~\cite{sun2019pullnet} trains a retrieval model composed of a LSTM and a graph neural network to retrieve a question-specific subgraph. NSM~\cite{he2021improving} and SR+NSM~\cite{zhang2022subgraph} propose a relation-path retrieval to retrieve subgraphs for multi-hop reasoning. EPR~\cite{ding2024enhancing} proposes an evidence pattern retrieval method to enhance retrieval-based methods in KGQA.

(3) LLM-based retrieval methods, including fine-tuned and prompting methods. Fine-tuned methods: UniKGQA~\cite{jiang2022unikgqa} unifies the graph retrieval and reasoning process into a single model with LLMs. DeCAF~\cite{yu2022decaf} combines semantic parsing and LLMs reasoning to jointly generate answers, which also reach salient performance on KGQA tasks. RoG~\cite{luo2023reasoning} proposes a planning-retrieval-reasoning framework that retrieves reasoning paths from KGs to guide LLMs in conducting faithful reasoning. GNN-RAG~\cite{mavromatis2024gnn} adopts a lightweight graph neural network to effectively retrieve from KGs. RnG-KBQA~\cite{ye2021rng} first uses BERT to rank a set of enumerated candidate programs (up to a limited complexity), and then uses T5 to edit the top programs into more complex programs. TIARA~\cite{shu2022tiara} first uses BERT to retrieve a set of schema items, which are further used as the input, together with the question, to T5 for plan generation. They also apply constrained decoding but only for grammatically. FC-KBQA~\cite{zhang2023fc} proposes a fine-to-coarse composition framework to avoid knowledge entanglement and guarantee both generalization ability and logical interpretability. FlexKBQA~\cite{li2024flexkbqa} is a flexible KGQA framework with LLMs. It can utilize a limited set of annotated data to build KGQA for different KGs and query languages. GAIN~\cite{shu2024distribution} pays attention to the robustness of KGQA models. It proposes a data augmentation method to alleviate this problem and further evaluates the distribution shifts including from different aspects. Prompting methods: StructGPT~\cite{jiang2023structgpt} defines the interface of KG data to implement knowledge access and filtering with finite quantity, and leverage the LLM to infer the answer or subsequent planning repeatedly. Interactive KBQA~\cite{xiong2024interactive} interacts with KGs directly and then generates logical forms. The interactions are under three designed universal APIs for KGs. ToG~\cite{sun2023think} iteratively retrieves relevant triplets from KGs and employs the LLM to assess whether the reasoning paths in beam search are sufficient for answering the question. PoG~\cite{chen2024plan} proposes a self-correcting adaptive planning paradigm for KG-augmented LLM, which effectively augmenting LLM’s reasoning ability and efficiency. ToG2.0~\cite{ma2024think} is a hybrid RAG framework that iteratively retrieves information from both unstructured and structured knowledge sources in a tight-coupling manner. KB-BINDER~\cite{li2023few} is developed to challenge the heterogeneity of items from different KGs. It enables few-shot in-context learning over KGQA tasks.

\noindent \textbf{Implementation Details.} We try four LLMs as the policy model in experiments: ChatGPT, GPT-4o-mini, GPT-4 and Qwen2.5-14b. We use OpenAI API to call ChatGPT (GPT-3.5-turbo-0806), GPT-4o-mini and GPT-4. Qwen2.5-instruct-14b runs with 8 A40-40G without quantization, where the temperature parameter is set to 0.7. The maximum token length for the generation is set to 256. We set the maximum flexible width of tree $b$ to 7 in all the datasets, and set iteration $H$ in reasoning tree construction to 24 in the WebQSP, CWQ and WebQuestions. In GrailQA, $H$ is set to 18. The maximum depth for judging leaf nodes is set to 5. The hyper-parameter $\alpha$ in evaluation is set to 0.33. The filtered $K$ weighted reasoning paths is 10 in all datasets. The Freebase is used as KG for CWQ, WebQSP, GrailQA and Webquestions.

\begin{table}[h]
\centering
\caption{Performance comparison on the WebQSP and CWQ. The best results are in bold and the second best results are underlined.}
\begin{tabular}{ccc}
\toprule
\multicolumn{1}{l}{Methods}                &WebQSP(\textbf{EM}$\uparrow$)   &CWQ(\textbf{EM}$\uparrow$)    \\ \midrule
\multicolumn{3}{c}{LLM-Only Methods}                 \\ \midrule
\multicolumn{1}{l}{Llama-3.1-8b}             &55.5         &28.1             \\
\multicolumn{1}{l}{Qwen2.5-14b}            & 59.0          &39.8         \\
\multicolumn{1}{l}{ChatGPT}                  & 62.6          & 38.8        \\
\multicolumn{1}{l}{GPT-4}                  & 67.3          & 46.0        \\
\multicolumn{1}{l}{GPT-4o-mini}            & 63.8          & 45.4        \\ \midrule
\multicolumn{3}{c}{Graph-based Retrieval Methods} \\ \midrule
\multicolumn{1}{l}{GraftNet}               & 66.7      & 36.8    \\
\multicolumn{1}{l}{PullNet}                & 68.1      & 45.9    \\
\multicolumn{1}{l}{NSM}                & 68.7      & 48.6    \\
\multicolumn{1}{l}{SR+NSM}                 & 68.9      & 50.2    \\
\multicolumn{1}{l}{ReaRev}                 & 76.4      & 52.9    \\
\multicolumn{1}{l}{ERM+NSM}                & 71.2      & 60.6    \\ \midrule
\multicolumn{3}{c}{LLM-based Retrieval Methods} \\ \midrule
\multicolumn{1}{l}{UniKGQA}                & 79.1      & 51.2    \\
\multicolumn{1}{l}{DeCAF}                  & 82.1      & 70.4    \\
\multicolumn{1}{l}{RoG}                   & 85.7      & 62.6    \\
\multicolumn{1}{l}{GNN-RAG}                & 85.7      & 66.8    \\
\multicolumn{1}{l}{StructGPT}              & 72.6      & 54.3    \\
\multicolumn{1}{l}{Interactive KBQA}       & 72.5      & 59.2    \\
\multicolumn{1}{l}{ToG w/ChatGPT}            & 76.2      & 57.1    \\
\multicolumn{1}{l}{ToG w/GPT-4}            & 82.6      & 68.5    \\
\multicolumn{1}{l}{ToG2.0}                 & 81.1      & -       \\
\multicolumn{1}{l}{EPREM}            & \underline{88.8}      & 66.2    \\
\multicolumn{1}{l}{PoG w/ChatGPT}            & 82.0      & 63.2    \\
\multicolumn{1}{l}{PoG w/GPT4}             &  87.3      & \underline{75.0}    \\ \midrule
\multicolumn{1}{l}{RTSoG w/ChatGPT}         &90.5 (+1.7\%)          &79.3 (+4.3\%)        \\
\multicolumn{1}{l}{RTSoG w/Qwen2.5-14b}     &92.2 (+3.4\%)         &78.9 (+3.9\%)        \\
\multicolumn{1}{l}{RTSoG w/GPT-4o-mini}     &93.1 (+4.3\%)          & 81.6 (+6.6\%)       \\
\multicolumn{1}{l}{RTSoG w/GPT-4}           &\textbf{94.3} \textbf{(+5.5\%)}           & \textbf{82.8} \textbf{(+7.8\%)}       \\ \bottomrule
\end{tabular}
\label{tab2}
\end{table}

\begin{table*}[h]
\centering
\caption{Performance comparison on the GrailQA and WebQuestions. The best results are in bold and the second best results are underlined.}
\begin{tabular}{cccccc}
\toprule
\multirow{2}{*}{Methods} & \multicolumn{4}{c}{GrailQA (\textbf{EM}$\uparrow$)}                & \multicolumn{1}{l}{\multirow{2}{*}{WebQuestions (\textbf{EM}$\uparrow$)}} \\ \cmidrule{2-5}
                         & Overall & I.I.D & Compositional & Zero-shot & \multicolumn{1}{l}{}                              \\ \midrule
\multicolumn{6}{c}{LLM-Only Methods}                                                                                              \\ \midrule
\multicolumn{1}{l}{ChatGPT}                  & 33.6        & 34.6      & 28.5              & 27.9         &48.5                                                   \\
\multicolumn{1}{l}{Qwen2.5-14b}                  & 31.6        & 32.4      & 36.0              & 34.9         &48.5                                                   \\
\multicolumn{1}{l}{GPT-4o-mini}              & 36.7        & 36.8      & 32.2              & 31.8         &52.3                                                   \\
\multicolumn{1}{l}{GPT4}                  & 39.7        & 40.7      & 34.7              & 33.5         &54.7                                                  \\ \midrule
\multicolumn{6}{c}{LLM-based Retrieval Methods (Fine-Tuned)}                                                                           \\ \midrule
\multicolumn{1}{l}{RnG-KBQA}                 & 68.8    & 86.2  & 63.8          & 63.0     & -                                                 \\
\multicolumn{1}{l}{TIARA}                    & 73.0    & 87.8  & 69.2          & 68.0     & -                                                 \\
\multicolumn{1}{l}{FC-KBQA}                  & 73.2    & 88.5  & 70.0          & 67.6     & -                                                 \\
\multicolumn{1}{l}{FlexKBQA}                 & 62.8    & 71.3  & 59.1          & 60.6     & -                                                 \\
\multicolumn{1}{l}{GAIN}                     & 76.3    & \underline{88.5}  & \underline{73.7}          & 71.8     & -                                                 \\ \midrule
\multicolumn{6}{c}{LLM-based Retrieval Methods (Prompting)}                                                                            \\ \midrule
\multicolumn{1}{l}{KB-BINDER}                & 50.6    & -     & -             & -        & -                                                 \\
\multicolumn{1}{l}{ToG w/ChatGPT}            & 68.7    & 70.1  & 56.1          & 72.7     & 54.5                                                 \\
\multicolumn{1}{l}{ToG w/GPT-4}              & 81.4    & 79.4  & 67.3          & 86.5     & 57.9                                              \\
\multicolumn{1}{l}{PoG w/ChatGPT}            & 76.5    & 76.3  & 62.1          & 81.7     & 60.6                                                \\
\multicolumn{1}{l}{PoG w/GPT4}               & \underline{84.7}    & 87.9  & 69.7          & \underline{88.6}     & \underline{67.1}                                              \\ \midrule
\multicolumn{1}{l}{RTSoG w/ChatGPT}           & 91.5(+6.8\%)    & 91.6(+3.1\%)   & 81.3(+7.6\%)         & 95.0(+6.4\%)     & 82.2(+15.1\%)                                                  \\
\multicolumn{1}{l}{RTSoG w/Qwen2.5-14b}       & 91.8(+7.1\%)    & 92.5(+4.0\%)      & 81.3(+7.6\%)              & 95.2(+6.6\%)         & 81.7(+14.6\%)                                                   \\
\multicolumn{1}{l}{RTSoG w/GPT-4o-mini}       & 92.8(+8.1\%)    & 93.7(+5.2\%)      & 83.3(+9.6\%)               & 95.7(+7.1\%)         & 82.7(+15.6\%)                                                  \\
\multicolumn{1}{l}{RTSoG w/GPT-4}             & \textbf{93.4}\textbf{(+8.7\%)}    & \textbf{94.5}\textbf{(+8.0\%)}      & \textbf{84.8}\textbf{(+11.1\%)}              & \textbf{95.9}\textbf{(+7.3\%)}        & \textbf{83.5}\textbf{(+16.4\%)}                                                  \\ \bottomrule
\end{tabular}
\label{tab3}
\end{table*}

\subsection{Main Results (EQ1)}
Table~\ref{tab2} and Table~\ref{tab3} show the result of RTSoG and other baselines across four representative KGQA datasets (WebQSP, CWQ, GrailQA and WebQuestions). Overall, regardless of which LLM is used as the policy model, RTSoG achieves the best performance across all four datasets. Specifically, we can make the following observations. 

First, in the WebQSP and CWQ datasets, compared to traditional Graph-based retrieval methods and LLM-only methods, the LLM-based retrieval methods show significant improvements. Notably, RTSoG (with GPT-4) achieved the best results, surpassing the ReaRev by 17.9\% and 29.9\% on the WebQSP and CWQ, respectively. Moreover, RTSoG consistently outperformed other LLM-based retrieval methods across different LLMs used as policy models. Compared to the current best PoG (with GPT-4), RTSoG showed an increases of 7.0\% and 7.8\% on WebQSP and CWQ under the same conditions. It is noteworthy that even with a smaller open-source model (e.g., qwen2.5-14b), RTSoG achieves results comparable to those of larger closed-source models (e.g., GPT-4 and ChatGPT). Specifically, RTSoG (with qwen2.5-14b) achieved improvements of 4.9\% and 3.9\% on two datasets compared to the PoG (with GPT-4). 

Secondly, across both GrailQA and WebQuestion datasets, RTSoG consistently outperformed both fine-tuned LLM-based retrieval methods and prompted LLM-based retrieval methods, regardless of which LLMs are employed as policy models or value models. Notably, on GrailQA, RTSoG achieves an 8.7\% performance improvement over PoG when using the same large language model (GPT-4). For more complex compositional problems, RTSoG demonstrates even more substantial enhancements, reaching a 15.1\% improvement. 
This indicates that through intent decomposition and reasoning tree construction, RTSoG possesses superior capabilities in problem comprehension and relevant information retrieval. Furthermore, compared to fine-tuning approaches, RTSoG shows more pronounced improvements, delivering an average 17.1\% gain over GAIN.

Finally, as a training-free prompting method, RTSoG exhibits relatively divergent outcomes when different LLMs are adopted as policy models. Generally, when employing larger and more advanced models such as GPT-4 and GPT-4o-mini, RTSoG typically shows marginal superiority over ChatGPT and qwen2.5-14b (averaging a 2\% improvement). However, given that OpenAI's models are not open-source and considering cost efficiency, practitioners may opt for smaller yet open-source models (e.g., qwen2.5-14b) as both policy and value models in practical applications.

\begin{figure*}[htbp]
\centering
\subfloat[]
{\includegraphics[height=0.16\textwidth]{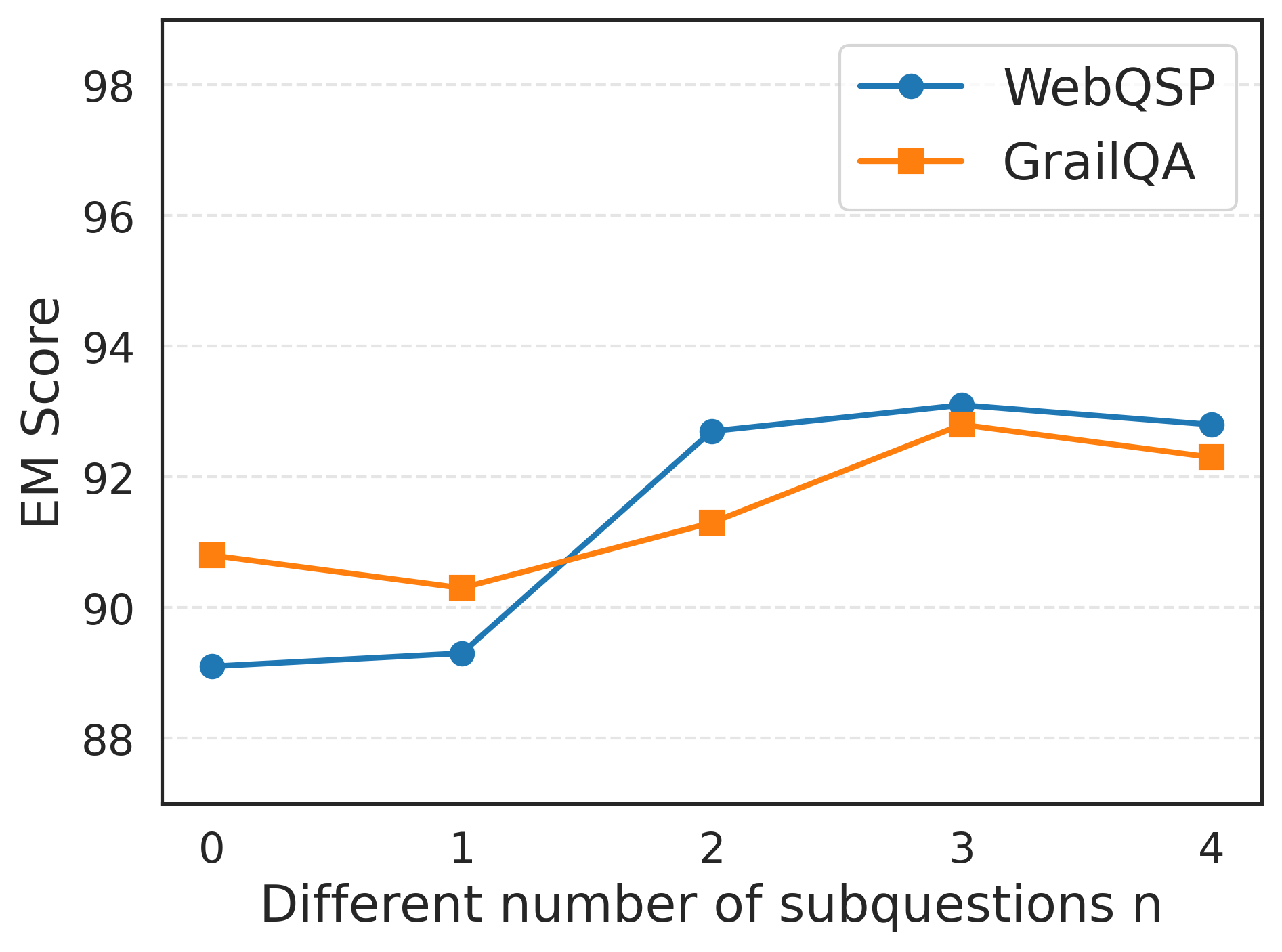}}
\subfloat[]
{\includegraphics[height=0.16\textwidth]{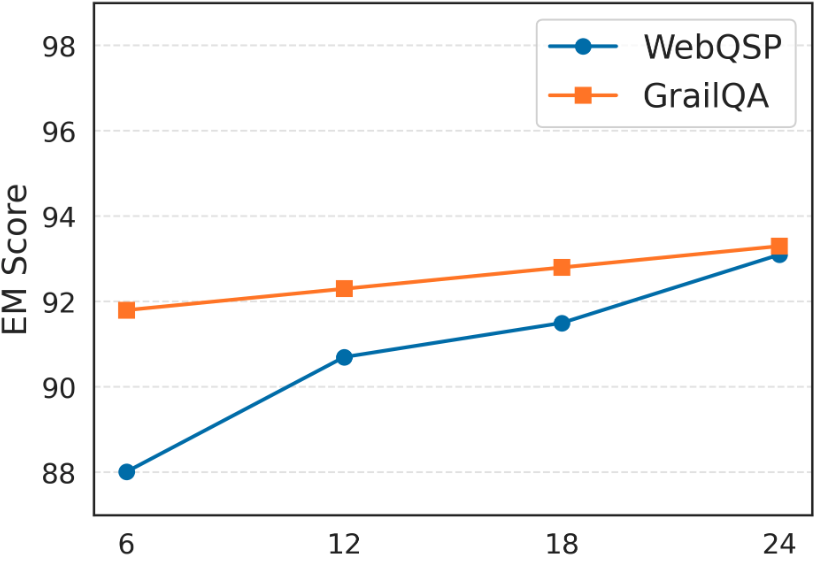}}
\subfloat[]
{\includegraphics[height=0.16\textwidth]{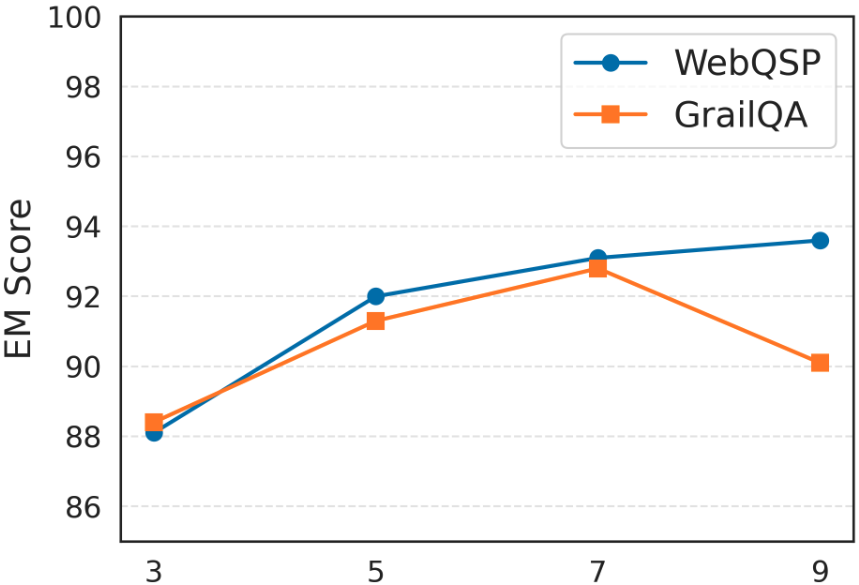}}
\subfloat[]
{\includegraphics[height=0.16\textwidth]{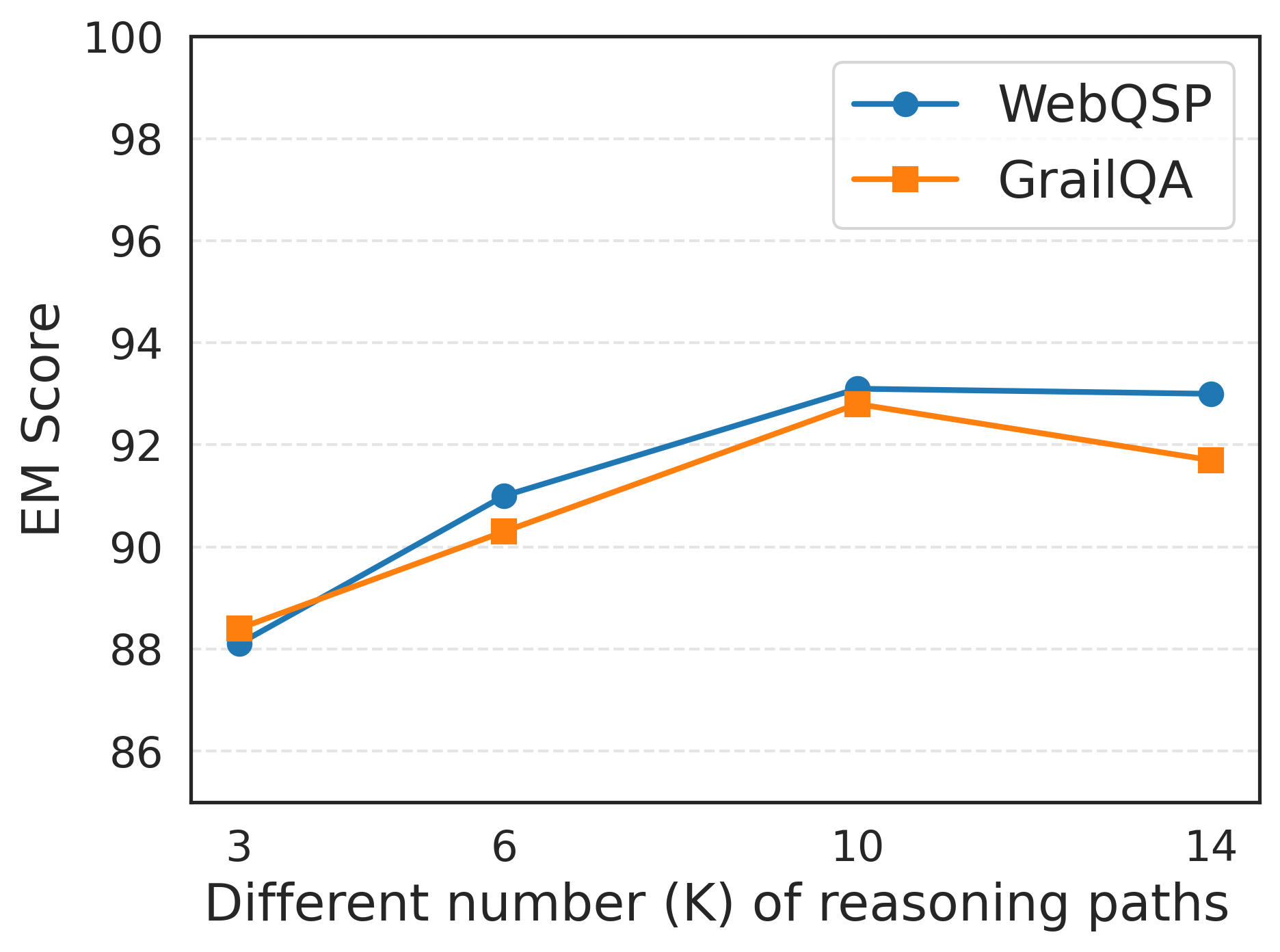}}
\label{fig2}
\caption{The impact of four important hyper-parameters: the number of subquestions $n$, the number of iterations $H$, the flexible width of the tree $b$ and the number of paths $K$ in $\mathcal{S}$.}
\label{p2}
\end{figure*}

\subsection{Sensitivity Analysis (EQ2)}
In this subsection, we conduct several experiments to analyze the impact of four important hyper-parameters: the number of subquestions $n$, the number of iterations $H$, the flexible width of the tree $b$ and the number of paths $K$ in $\mathcal{S}$. We conduct these experiments on WebQSP and GrailQA, using GPT-4o-mini as the policy model. First, we evaluate the impact of the different number of subquestions $n$. As shown in Fig~\ref{p2}(a), when RTSoG does not perform question decomposition, the final performance is significantly lower than when question decomposition is applied. This also indicates that question decomposition can effectively reduce the difficulty of the question and assist the subsequent SC-MCTS in exploring reasoning paths. However, blindly increasing the number of question decompositions does not lead to continuous performance improvement, as the key points of the question can only be broken down to a limited extent. When the number of subquestion is 3, the RTSoG demonstrates higher performance on both datasets.

Second, we vary the number of iterations $H$ (6, 12, 18, 24) during the construction of the reasoning tree. As shown in Fig~\ref{p2}(b), when $k$ is fixed, the EM scores of the final prediction gradually increase with the number of iterations. This is because a higher number of iterations allows for the exploration of more relevant knowledge in the knowledge graph, thereby increasing the number of correct weighted paths. However, balancing efficiency and final results, we choose 24 iterations for WebQSP and 18 iterations for GrailQA.

Next, we change the flexible width of the tree $b$ (3, 5, 7, 9) during the expansion step. As seen in Fig~\ref{p2}(c), when $H$ is fixed, the EM scores of the final prediction first gradually increase with the width of the reasoning tree during the expansion step, then tend to stabilize or decrease. In GrailQA, it can be observed that when $b$ rises to 9, the EM scores even decrease. This phenomenon occurs because as $b$ increases, the model can explore more correct reasoning paths, but when the width of the expansion tree is too large, it introduces noise paths, which can degrade the performance of the model. Therefore, considering both efficiency and result accuracy, $b$ is set to 7 for both WebQSP and GrailQA.

Finally, we change the number of reasoning paths in stack $\mathcal{S}$. The result in Fig~\ref{p2}(d). From these results, we can observe that the performance declines when too few reasoning paths are retained. This may be because the value model can not generate the accurate reward of paths, leading to the filtering out of some important reasoning paths.
However, if the number of selected paths is too large, it introduces noisy reasoning paths, which negatively impacts the final answers. When the number of reasoning paths is set to 10, the model achieves optimal performance across multiple datasets.

\begin{figure}[htbp]
\centering
\subfloat[]
{\includegraphics[height=0.17\textwidth]{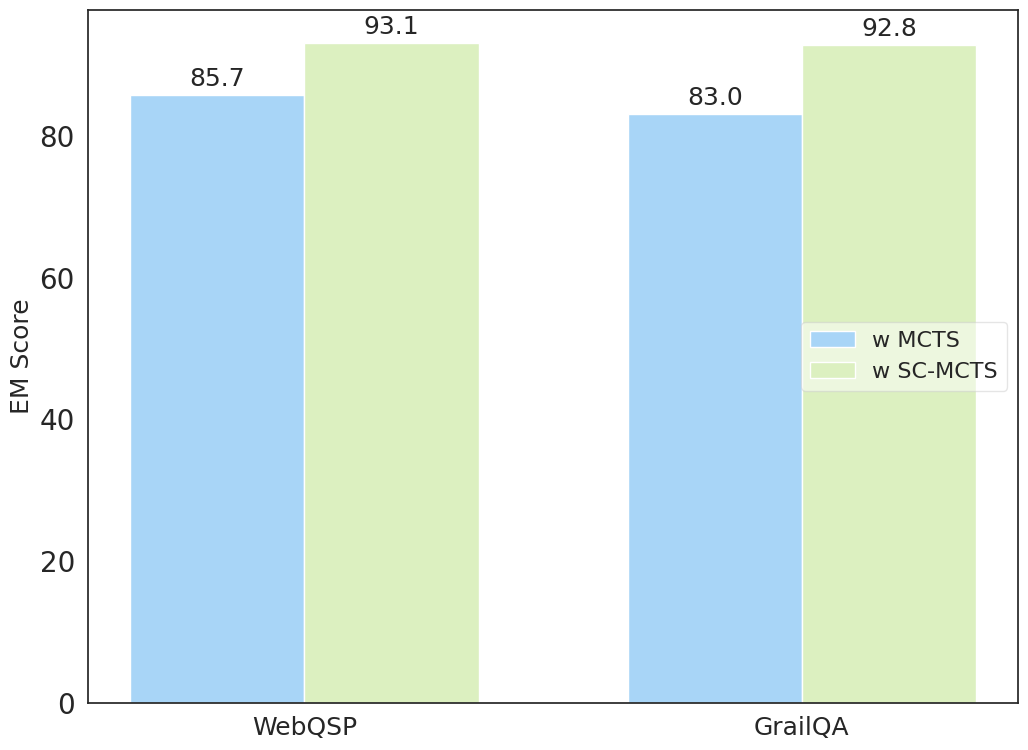}}
\subfloat[]
{\includegraphics[height=0.17\textwidth]{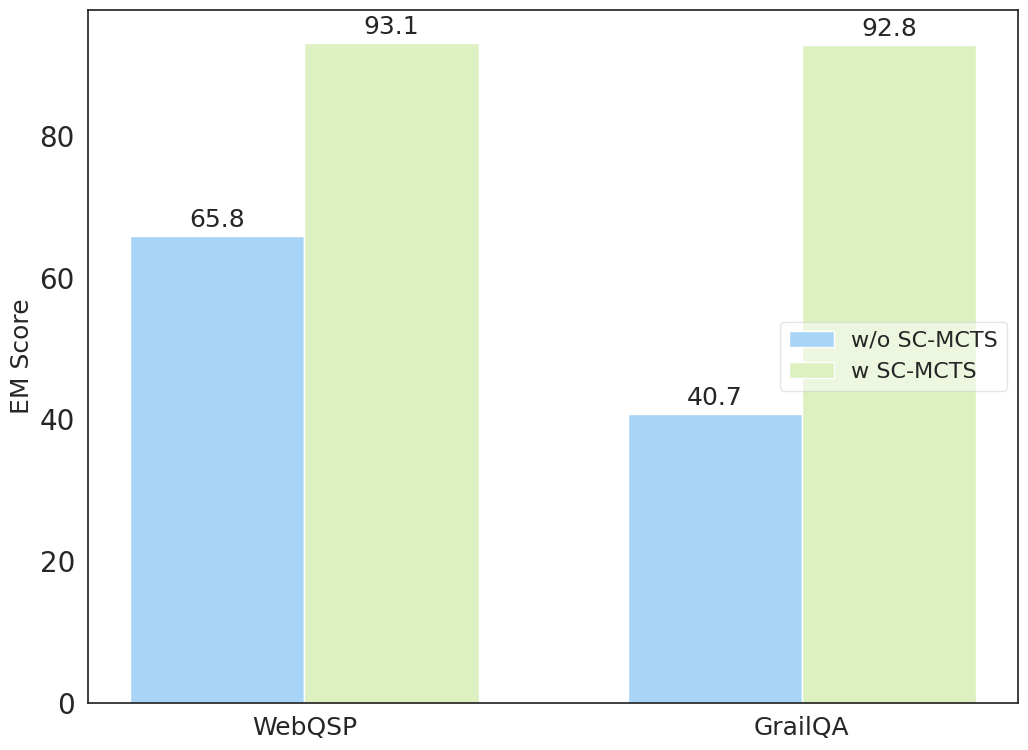}}
\label{fig2}
\caption{Ablation study on the self-critic mechanism and SC-MCTS in RTSoG.}
\label{p3}
\end{figure}

\subsection{Ablation Study (EQ3)} 
In this subsection, we conduct two ablation studies to analyze the SC-MCTS in the weighted reasoning paths retrieval.
First, we replaced the SC-MCTS in RTSoG with MCTS without the self-critic mechanism. As shown in Fig~\ref{p3}(a), under the same number of iteration $H$ and tree width $b$, SC-MCTS achieved consistent improvements over MCTS on both datasets: 7.4\% on WebQSP and 9.8\% on CWQ. This improvement stems from the self-critic mechanism, which allows timely termination of node expansion when further exploration is unnecessary. This not only reduces the noisy paths but also minimizes redundant computations. 

Second, we removed the SC-MCTS in RTSoG and directly utilized the decomposed sub-questions to guide the LLM to generate answers. According to the results in Fig~\ref{p3}(b), it is evident that omitting the weighted paths retrieval stage achieved by SC-MCTS significantly reduces the model's final performance, with a decrease of 27.3\% on WebQSP and 52.1\% on GrailQA. This result clearly demonstrates that the SC-MCTS effectively retrieves knowledge relevant to the questions from the knowledge graph, thereby better assisting the LLM in generating accurate answers. Additionally, it helps mitigate potential issues such as out-of-date knowledge in the LLM.

\begin{table}[]
\centering
\caption{Ablation study on the reasoning path stack in answer generation.}
\begin{tabular}{c|cc}
\toprule
\multirow{2}{*}{\textbf{Methods}}                                       & \multicolumn{2}{c}{\textbf{Datasets}} \\ \cmidrule{2-3} 
                                                                        & WebQSP            & GrailQA           \\ \midrule
RTSoG  \\ w/ Reasoning Path Stack                   & \textbf{93.1}     & \textbf{92.8}     \\ \midrule
\begin{tabular}[c]{@{}c@{}}RTSoG \\ w/o Reasoning Path Stack\end{tabular} & 87.9              &86.7                   \\ \bottomrule
\end{tabular}
\label{tab4}
\end{table}
Finally, to analyze the importance of the paths stack in answer generation, we directly use the weighted reasoning paths for answer generation, without employing the reasoning path stack. As shown in the table~\ref{tab4}, it indicates that using the reasoning path stack improves the model's final prediction EM score, with an increase of 5.2\% on WebQSP and 7.2\% on GrailQA. This finding suggests that when a series of weighted reasoning paths are obtained, fully considering the different weights of the paths can effectively eliminate some noisy paths, and improve the performance of the model.

\begin{table}[]
\caption{The number of LLM calls per question}
\centering
\begin{tabular}{ccccc}
\toprule
\textbf{Methods}   & ToG    &TOG2.0    & PoG      & RTSoG \\ \midrule
\textbf{LLM Calls} & $H\times D$ & $H\times D$    & $H\times d_1$ & $2\times H \times b$  \\ \bottomrule
\end{tabular}
\label{tab5}
\end{table}

\subsection{Efficiency Analysis (EQ4)} 
In this subsection, we will discuss the efficiency of the model. Assuming the number of iterations of all the methods is $H$, and the beam search width in ToG is $D$, the number of LLM calls for the existing methods is shown in the table~\ref{tab5}. 
$d_1$ is the exploration width of PoG, $k$ is the flexible width of the reasoning tree in RTSoG, and $d_1 < D$, $k < D$. Although RTSoG requires LLM calls during both the expansion and evaluation steps, the number of LLM calls is still comparable to other methods. It is noteworthy that RTSoG achieves better results when using smaller open-source models like Qwen2.5-14b compared to other baselines on closed-source large models like GPT-4. This allows RTSoG to avoid the costly API call expenses, resulting in a lower cost for practical applications.

\begin{figure}[htbp]
\centering 
{\includegraphics[height=0.52\textwidth]{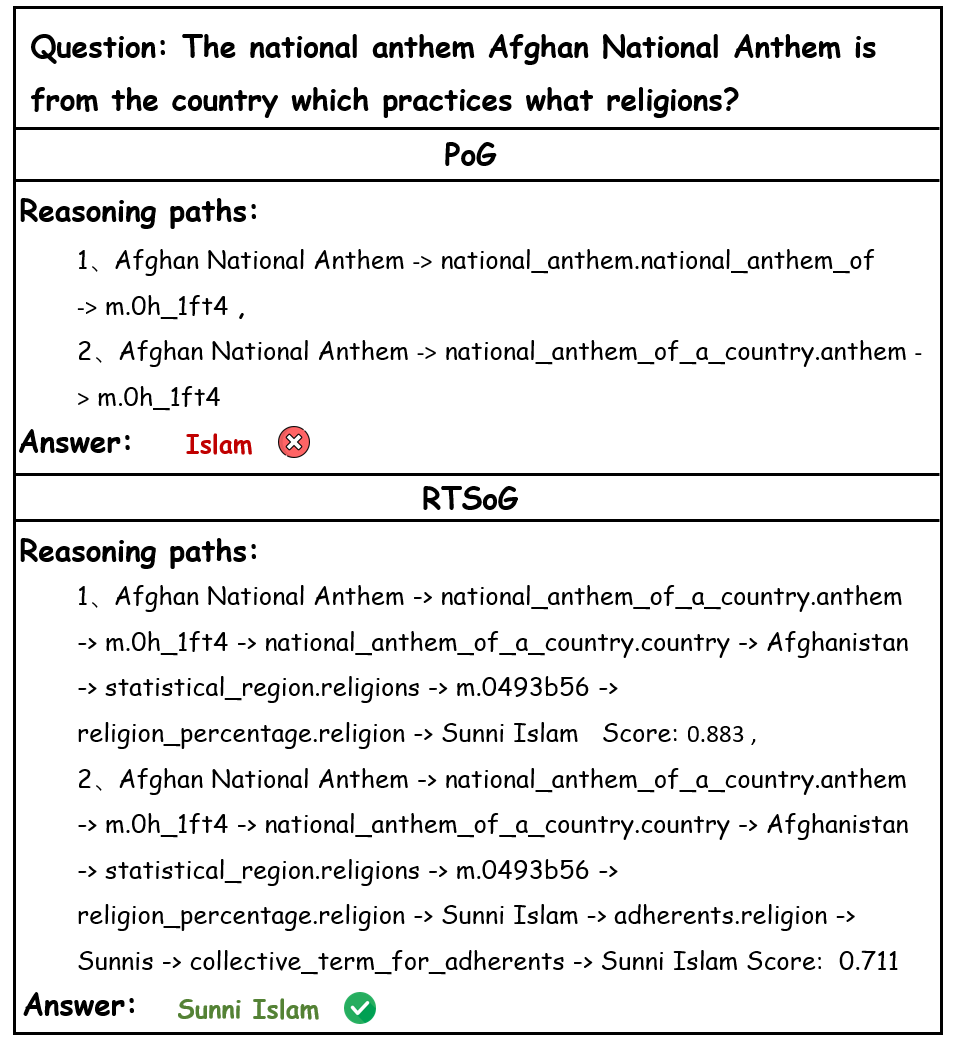}}
\caption{A typical case to analyze the difference between PoG and RTSoG in exploring reasoning paths.} 
\label{fig4}
\end{figure} 

\begin{figure}[htbp]
\centering 
{\includegraphics[height=0.5\textwidth]{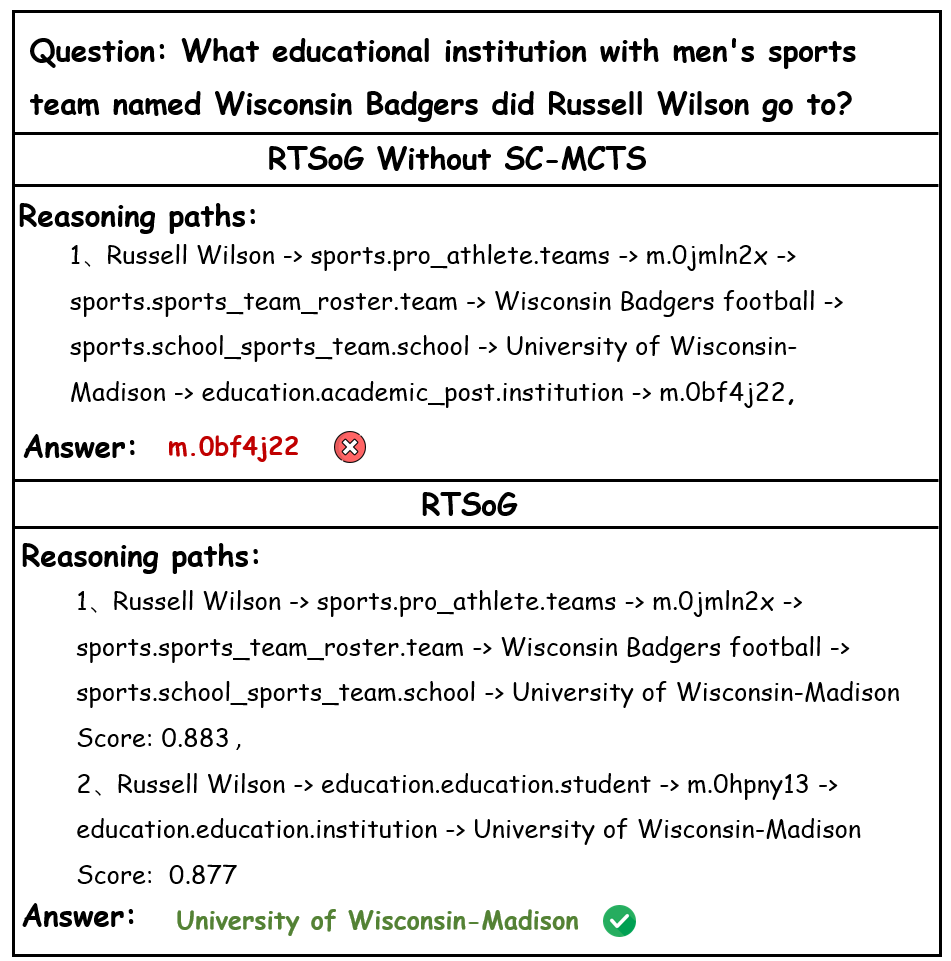}}
\caption{A typical case to to analyze the impact of SC-MCTS in weighted reasoning paths retrieval.} 
\label{fig5}
\end{figure} 

\subsection{Case Study (EQ5)}
In this subsection, we will conduct two case studies to analyze how different methods explore reasoning paths and the impact of SC-MCTS on RTSoG's reasoning results. 

First, to analyze the difference between PoG and RTSoG in exploring reasoning paths, fig~\ref{fig4} shows a typical case in the CWQ dataset. To answer the question ``The national anthem Afghan National Anthem is from the country which practices what religions?", it can be observed that the reasoning paths retrieved by PoG are relatively short and do not provide critical information for the final reasoning. The mid-form entities ``m.0493b56" lack semantic meaning. Therefore, the wrong answer ``Islam" generated by PoG relies on the internal knowledge of the LLM, and the retrieved information does not contribute to enhanced reasoning. However, the weighted reasoning paths retrieved by RTSoG indicate that the ``Afghan National Anthem" is the national anthem of ``Afghanistan", and within Afghanistan's religious distribution, Sunni Islam is one of the predominant religions. Moreover, this reasoning path has a high score, providing strong support for the final answer generation. 

Second, to analyze the impact of SC-MCTS in weighted reasoning paths retrieval, fig~\ref{fig5} shows a case in the CWQ. The question is ``What educational institution with men's sports team named Wisconsin Badgers did Russell Wilson go to?". When we replaced the SC-MCTS with MCTS in RTSoG, the final explored reasoning path is shown in the upper part of the fig, while the lower part displays the high-reward reasoning paths discovered by the RTSoG. The path explored by RTSoG without SC-MCTS shows that after reaching the answer University of Wisconsin-Madison, the exploration does not stop, ultimately leading the reasoning path to an incorrect answer. This also demonstrates the importance of the self-critic mechanism in SC-MCTS, which enables the timely termination of the exploration, which not only reduces the noisy paths but also minimizes redundant computations. 

\section{Conclusion and future work}
In this paper, we propose a novel framework named Reward-guided Tree Search on Graph (RTSoG) for KGQA tasks, which uses the Monte Carlo Tree Search (MCTS) to better balance the exploration and exploitation of reasoning paths in KGs. Specifically, it first decomposes the original questions into a series of simpler and well-defined sub-questions. Second, RTSoG iteratively performs the Self-Critic Monte Carlo Tree Search (SC-MCTS) on knowledge graphs to retrieve weighted reasoning paths that support reasoning the answers. Finally, it pushes the reasoning paths into the stack in order of their weight, to generate the final answers. Extensive experiments on four datasets demonstrate the effectiveness of RTSoG. Typically, it achieves 8.7\% and 7.0\% improvement over the state-of-the-art method on the GrailQA and the WebQSP respectively.


\bibliography{tkde}
\bibliographystyle{IEEEtran}

\vfill

\end{document}